\documentclass[runningheads]{llncs}

\usepackage{eccv}

\usepackage{eccvabbrv}

\usepackage{graphicx}
\usepackage{booktabs}

\usepackage[accsupp]{axessibility}  

\usepackage{hyperref}

\usepackage{orcidlink}

\begin{document}

\title{MMA-MRNNet: Harnessing Multiple Models of Affect and Dynamic Masked RNN for Precise Facial Expression Intensity Estimation} 

\titlerunning{MMA-MRNNet for Facial Expression Intensity Estimation}

\author{Dimitrios Kollias\inst{1} 
\and
Andreas Psaroudakis\inst{2} 
\and
Anastasios Arsenos\inst{2} 
\and
Paraskevi Theofilou\inst{2}
\and
Chunchang Shao\inst{1}
\and
Guanyu Hu\inst{1}  
\and
Ioannis Patras\inst{1}  
}

\authorrunning{D.~Kollias et al.}

\institute{Queen Mary University of London, UK \\
\and
National Technical University of Athens, Greece\\
\email{d.kollias@qmul.ac.uk}
}

\maketitle

\begin{abstract}
 %
This paper presents MMA-MRNNet, a novel deep learning architecture for dynamic multi-output Facial Expression Intensity Estimation (FEIE) from video data. Traditional approaches to this task often rely on complex 3-D CNNs, which require extensive pre-training and assume that facial expressions are uniformly distributed across all frames of a video. These methods struggle to handle videos of varying lengths, often resorting to ad-hoc strategies that either discard valuable information or introduce bias. MMA-MRNNet addresses these challenges through a two-stage process. First, the Multiple Models of Affect (MMA) extractor component is a Multi-Task Learning CNN that concurrently estimates valence-arousal, recognizes basic facial expressions, and detects action units in each frame. These representations are then processed by a Masked RNN component, which captures temporal dependencies and dynamically updates weights according to the true length of the input video, ensuring that only the most relevant features are used for the final prediction. The proposed unimodal non-ensemble learning MMA-MRNNet was evaluated on the Hume-Reaction dataset and demonstrated significantly superior performance, surpassing state-of-the-art methods by a wide margin, regardless of whether they were unimodal, multimodal, or ensemble approaches. Finally, we demonstrated the effectiveness of the MMA component of our proposed method across multiple in-the-wild datasets, where it consistently outperformed all state-of-the-art methods across various metrics.

\end{abstract}

\section{Introduction}
\label{sec:intro}

Human emotions are complex, conscious experiences that profoundly influence behavior and can be expressed in various forms. These emotions are pivotal in psychological processes and significantly impact human actions. The advent of Artificial Intelligence (AI) and Deep Learning (DL) has driven the development of intelligent systems capable of recognizing and interpreting human emotions. Psychologists have proposed multiple descriptors to quantify and categorize emotional states: sparse descriptors like facial action units (AUs), which capture specific facial muscle activations \cite{ekman1978facial}; continuous descriptors such as valence and arousal, where valence indicates the positivity or negativity of the emotion, and arousal reflects the level of activation or passivity \cite{plutchik1982psychoevolutionary}; and discrete class descriptors like the six basic expressions (anger, disgust, fear, happiness, sadness, surprise) and the neutral state \cite{ekman2002facial}. This paper focuses on dynamic multi-output Facial Expression Intensity Estimation (FEIE), specifically targeting the intensity estimation of expressions such as Adoration, Amusement, Anxiety, Disgust, Empathic-Pain, Fear, and Surprise.

In this paper, we introduce our approach MMA-MRNNet, a novel deep learning architecture designed to tackle the complexities of FEIE in scenarios where video-level annotations (i.e., there exists one annotation for the whole video) are provided rather than frame-level annotations. The key challenges addressed by MMA-MRNNet include handling the variability in video lengths and accurately aggregating temporal information across frames to make a robust final prediction. 

Traditional approaches for processing 3-D signals, such as video data, typically employ 3-D CNNs that produce a single prediction per signal. However, these architectures are inherently complex, with a high number of parameters, and often require pre-training on large 3-D datasets to achieve satisfactory performance. Another common approach involves assigning the video-level label uniformly to each frame and then using CNN-RNN networks to train on these annotated frames. This approach assumes that the facial expression intensity is consistent across all frames, which may not be the case, as only a subset of frames might actually display the labeled intensity \cite{kollias2023deep,arsenos2023data,kollias2023ai,kollias2024domain,kollias2023facernet,kollias2024distribution,kollias2023btdnet,gerogiannis2024covid}.

Moreover, our approach addresses the challenge of variable-length input videos. Traditional methods often rely on ad-hoc strategies to manage varying numbers of frames, such as setting a fixed input length and either discarding excess frames (which risks losing critical information) or duplicating frames in shorter videos (which can bias the model towards repeated data). These strategies are not only suboptimal but also require empirical tuning for each specific dataset, limiting their generalizability and effectiveness.



MMA-MRNNet comprises two primary components: the Multiple Models of Affect (MMA) extractor and the Masked RNN and Routing Network (MRNN). The MMA component is a Multi-Task Learning (MTL) CNN that extracts affective representations from each frame by concurrently estimating valence-arousal (VA), recognizing the 7 basic expressions, and detecting multiple action units (AUs). To ensure the reliability and consistency of these representations, we introduce a novel loss function that incorporates prior knowledge of the relationships between different affective descriptors, mitigating issues like noisy gradients and poor convergence typically encountered in MTL settings.

The extracted representations are then passed to the MRNN component, which consists of an RNN designed to capture temporal dependencies across the sequence of frames. To handle the varying lengths of input videos, a Mask layer is employed within the MRNN. This layer dynamically selects relevant RNN outputs based on the actual number of frames in the video, allowing the model to adapt to variable input lengths without compromising the integrity of the temporal information. The selected features are then passed through fully connected layers to produce the final intensity estimation for the entire video.

To the best of our knowledge, MMA-MRNNet is the first architecture to leverage valence-arousal, AUs, and basic expressions as intermediate representations for the task of Facial Expression Intensity Estimation. This approach not only enhances the model's ability to capture the nuanced dynamics of emotional expressions but also provides a robust framework for handling real-world data with varying input conditions.

\section{Related Work}

\cite{10.1145/3136755.3143009} presented Supervised Scoring Ensemble (SSE) for emotion recognition. A new fusion structure is presented in which class-wise scoring activations at diverse complementary feature layers are concatenated and used as inputs for second-level supervision, acting as a deep feature ensemble within a single CNN architecture.
\cite{https://doi.org/10.48550/arxiv.2003.00832} proposed a deep Visual-Audio Attention Network (VAANet) for video emotion recognition; VAANet integrates spatial, channel-wise, and temporal attentions into a visual 3D CNN and temporal attentions into an audio 2D CNN. A polarity-consistent cross-entropy loss is proposed for guiding the attention generation, which is based on the polarity-emotion hierarchy constraint.
\cite{https://doi.org/10.48550/arxiv.2002.09023} constructed an A/V hybrid network to recognize human emotions. A VGG-Face (for extracting per-frame features) and LSTM (for correlating these features according to their temporal dependencies) architecture was used for the visual data.

\cite{li2023multimodal} was the winning method of the Emotional Reaction Intensity (ERI) Estimation Challenge of the 5th ABAW Challenge \cite{kollias2019expression,kollias2020analysing,kollias2021affect,kollias2021analysing,kollias2021distribution,kollias2020deep,zafeiriou1,kollias2017recognition,kollias20222abaw,kollias2022abaw,kollias2023abaw2,kollias2023deep2,kollias2023abaw,kollias2023abaww,kollias2023facernet,kollias2023multi,kollias20246th,kollias2024distribution,kollias2024domain,arsenos2023data,kollias2023ai}. This method consists of an audio
feature encoding module (based on DenseNet121 and DeepSpectrum), a visual feature encoding module (based on PosterV2-Vit), and an audio-visual modality interaction module. 
\cite{vaiani2022viper} proposed ViPER, a modality agnostic late fusion network that leverages a transformer-based model that combines video frames, audio recordings, and textual annotations for FEIE. 
\cite{yu2023exploring} proposed a dual-branch FEIE model; the one branch (composed of Temporal CNN and Transformer encoder) handles the visual modality and the other handles the audio one; modality dropout is added for A/V feature fusion.
\cite{qiu2023multi} achieved the 3rd place in the ERI challenge of the 5th ABAW; it proposed a methodology that involved extracting features from visual, audio, and text modalities using Vision Transformers, HuBERT, and DeBERTa. Temporal augmentation and SE blocks were applied to enhance temporal generalization \cite{arsenos2024uncertainty,arsenos4674579nefeli,miah2024can,arsenos2024commonn,karampinis2024ensuring} and contextual understanding. Features from each modality were then processed through contextual layers and fused using a late fusion strategy.
\cite{wang2023emotional} presented a methodology that involved extracting visual features from video frames using models like FAb-Net, EfficientNet, and DAN, which capture facial expressions and attributes. Audio features are obtained using Wav2Vec2 and VGGish models. The extracted features were then processed through a temporal convolutional network to capture local temporal information, followed by a Transformer Encoder to model long-range dependencies with dynamic attention.
\cite{zhang2023facial} presented a methodology that involved  extracting audio and visual features using state-of-the-art models and aligning these features to a common dimension using an Affine Module. The aligned features were then fused using a Multimodal Multi-Head Attention model.

\section{Methodology}

\textbf{Formulation } The input to our method is a video
consisting of multiple instances (i.e., videoframes), $\textbf{\textit{X}} = \{\textbf{x}_1,..., \textbf{x}_K\}$, with $\textbf{x}_K \in \Re^{H \times W \times 3} $. $K$ is the number of instances (frames), which varies for
different videos; $H$ and $W$ denote the height and width of the RGB images (frames). There is a video-level label $\textit{Y}$. We further assume the instances also have corresponding instance-level labels $\{\textbf{y}_1,..., \textbf{y}_K\}$, which are unknown during training; the instance-level labels (of all instances of the same video) do not necessary match the video-level label. There are $N$ such video-label pairs constituting the database $DB = \{\textbf{\textit{X}}_n, \textit{Y}_n\}^\textit{N}_{n=1}$. Our objective is to learn an optimal function for predicting the video-level label with the video's instances as input. To this end, our method should be able to: \\
1) effectively consider the fact that input videos have variable lengths (in other words, the method should tackle the fact that the total number of frames varies for different videos) \\
2) aggregate the information of instances $\{\textbf{\textit{x}}_k\}^K_{k=1}$
 to make the final decision.
A well-adopted aggregation method is the embedding-based approach which maps $\textbf{\textit{X}}$ to a video-level representation $\textbf{\textit{z}} \in \Re^F$  and use $\textbf{\textit{z}}$ to predict $\textit{Y}$.

Initially, all videos $\{\textbf{\textit{X}}_n\}^N_{n=1}$ are padded to a uniform length $t$, resulting in video sequences $\textbf{\textit{X}}_N = \{\textbf{x}_1,..., \textbf{x}_t\}$. 
Each video $\textbf{\textit{X}}$ is then processed by the Multiple Models of Affect (MMA) extractor component, which conducts local analysis on each 2-D frame, mapping $\textbf{\textit{X}}$ to a multiple affect-level representation matrix $\textbf{\textit{Z}} = \{\textbf{z}_1,..., \textbf{z}_t\} \in \Re^{d \times t} $. This matrix is subsequently passed to an RNN, positioned on top of the MMA component, to capture temporal dependencies across all  $\{\textbf{\textit{z}}_k\}^t_{k=1}$. The RNN transforms $\textbf{\textit{Z}}$
into an embedding matrix $\textbf{\textit{Z}}' = \{\textbf{z}'_1,..., \textbf{z}'_t\} \in \Re^{d' \times t}$, performing global analysis over the entire video.
The subsequent module aggregates the set of embeddings $\{\textbf{\textit{z}}'_k\}_{k=1}^t$ into a single video-level vector embedding
$\textbf{\textit{z}}' \in \Re^{d' \cdot t}$, which is then fed to a Mask layer.
The Mask layer dynamically selects embeddings based on the 'true' frame count of the video, accounting for the original number of frames prior to padding. This step is crucial because the video-level annotations imply that all frames collectively, rather than individually, carry important information for an accurate prediction.
The output of the Mask layer $\textbf{\textit{z}}'' \in \Re^{d' \cdot t}$ is then mapped to another embedding $\textbf{\textit{z}}''' \in \Re^{d''}$ using a feed forward layer. Finally, $\textbf{\textit{z}}'''$ is transformed into the logits $\textbf{\textit{u}}$ via a feed forward layer parameterized by $\textbf{\textit{W}}$ leading to the video-level classification: $ 
 \textbf{\textit{u}} =  \textbf{\textit{W}}^T \textbf{\textit{z}}'''$.

\begin{figure*}[h]
\centering
  \includegraphics[width=1.\textwidth]{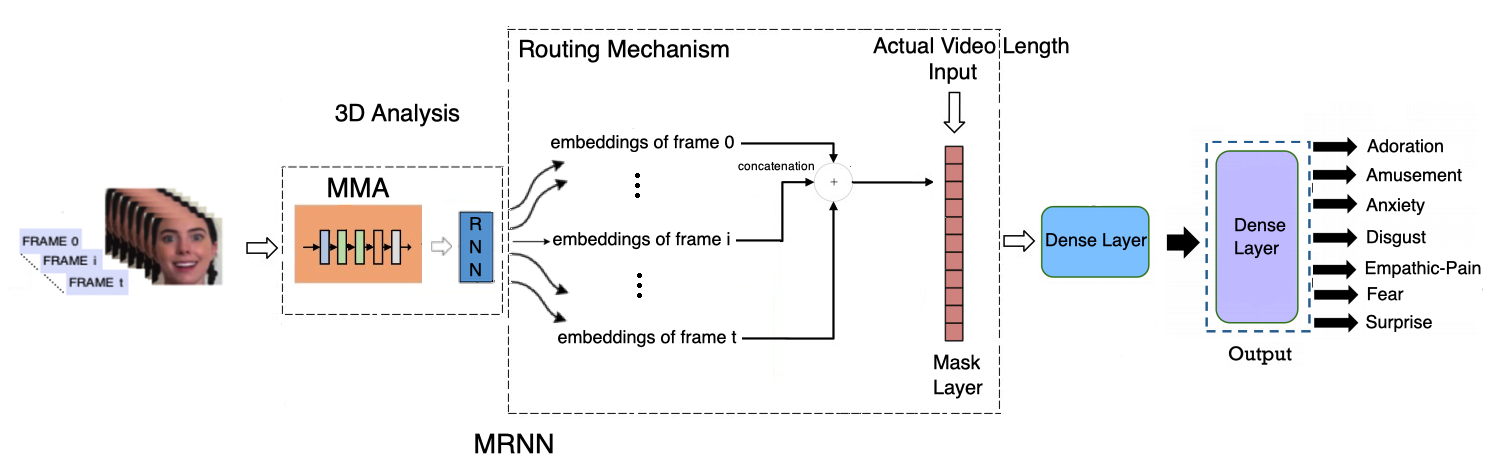}
  \caption{
  Overview of the proposed MMA-MRNNet for dynamic multi-output Facial Expression Intensity Estimation. MMA-MRNNet comprises two main components: the Multiple Models of Affect (MMA) extractor, which generates affective representations (valence-arousal, basic expressions, and action units) for each video frame, and the Masked  RNN and Routing (MRNN), which captures temporal dependencies and dynamically selects key features (and updates weights) according to the variable lengths of input videos. 
  }
  \label{methodology_iccv}
\end{figure*}

In the following, we further explain in more detail each component of our proposed method.
Fig. \ref{methodology_iccv} gives an overview of our proposed framework, MMA-MRNNet.

\subsection{MMA: Multiple Models of Affect extractor Component}

\begin{figure}[h]
\centering
  \includegraphics[height=7cm]{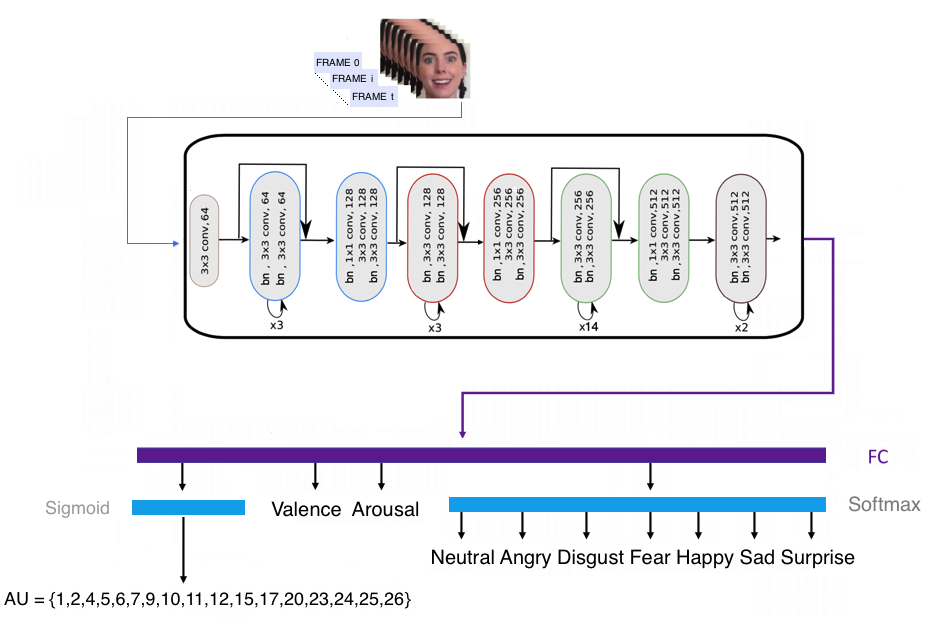}
  \caption{The Mulptiple Models of Affect extractor Component (MMA) that outputs for each frame the following emotional descriptors: valence and arousal, 17 action units and 7 basic expressions}
  \label{methodology_iccv2}
\end{figure}

The Multiple Models of Affect (MMA) extractor component processes an input video $\textbf{\textit{X}}$ by extracting affective representations from each frame  using three distinct models of affect. Specifically, the MMA is a Multi-Task Learning (MTL) CNN model that concurrently performs: (i) continuous affect estimation in terms of valence and arousal (VA); (ii) recognition of 7 basic facial expressions; and (iii) detection of 17 action units (AUs). The architecture of the MMA, illustrated in Fig. \ref{methodology_iccv2}, is structured around residual units, with 'bn' indicating batch normalization layers.
The model integrates the valence-arousal estimation, 7 basic expression recognition, and 17 AU detection tasks within the same embedding space derived from a shared feed-forward layer.
Consequently, the output of the MMA when processing $\textbf{\textit{X}}$ is a multiple affect-level representation matrix $\textbf{\textit{Z}} = \{\textbf{z}_1,..., \textbf{z}_t\} \in \Re^{26 \times t}$.

For training the MMA, we utilize multiple in-the-wild datasets, including Aff-Wild2 \cite{kollias2019deep,kollias2019expression,kollias2022abaw,kollias2023abaww,kollias2023multi,kollias20246th,kollias2024distribution,kolliasijcv,zafeiriou2017aff,kollias2020analysing,kollias2021distribution,kollias2021affect,kollias2019face,kollias2021analysing}, AffectNet \cite{mollahosseini2017affectnet}, and EmotioNet \cite{emotionet2016}, which are annotated for valence-arousal, 7 basic expressions, and 17 action units (these action units are an aggregate in all datasets).

Our recent studies \cite{hu2024bridging,hu2024rethinking} have highlighted challenges in the current evaluation of affect analysis methods, noting inconsistencies in database partitioning and evaluation practices that lead to biased and unfair comparisons. To address these issues, a unified protocol for database partitioning was proposed, ensuring fairness and comparability, while also accounting for subjects' demographic information. It was demonstrated that methods previously considered state-of-the-art on original partitions may not retain their performance under this new protocol. Consequently, in this current paper, we adopt this updated partitioning protocol.

A key challenge in utilizing these datasets is the non-overlapping nature of their task-specific annotations. For instance, EmotioNet only includes AU annotations, lacking valence-arousal and 7 basic expression labels. Training the MMA directly with these datasets using a combined loss function for all tasks would result in noisy gradients and poor convergence, as not all loss terms would be consistently contributing to the overall objective function. This can lead to issues typical of Multi-Task Learning (MTL), such as task imbalance, where one task may dominate training, or negative transfer, where the MTL model underperforms compared to single-task models \cite{kollias2024distribution,kollias2021distribution,kollias2019face}.


To address this issue, we generate AU pseudo-representations ($r_{AU}^\prime$) from the 7 basic expression representations ($r_{expr}$) produced by the MMA for each frame. This is achieved by leveraging the relationship between expressions and AUs, as defined in Table 1 of \cite{du2014compound}. The study by \cite{du2014compound} conducted a cognitive and psychological analysis of the associations between facial expressions and AU activations, summarizing the findings in a table that details the relatedness between expressions and their corresponding AUs. This table is presented in Table \ref{relationship} for reference. Prototypical AUs are those consistently identified as activated by all annotators, while observational AUs are those marked as activated by only a subset of annotators.

\begin{table*}[h]
\centering
\caption{
Relatedness of expressions \& AUs inferred from \cite{du2014compound}
}
\begin{tabular}{|l|c|c|}
\hline
Expression   & Prototypical AUs & Observational AUs \\
\hline\hline
happiness &  12, 25 & 6    \\
\hline
sadness &  4, 15 & 1,6 , 11, 17  \\
\hline
fear &  1, 4, 20, 25 &2, 5, 26  \\
\hline
anger &4, 7, 24 &10, 17, 23 \\
\hline
surprise &1, 2, 25, 26 &5 \\
\hline
disgust &9, 10, 17 & 4,   24 \\
\hline
\end{tabular}
\label{relationship}
\end{table*}

The AU pseudo-representations are modeled as a mixture over the basic expression categories:

\begin{equation}
    r_{AU}^\prime = \sum\nolimits_{expr}  r_{expr} \cdot r_{{AU}|{expr}}
\label{eq:distr1}
\end{equation}

\noindent where $r_{{AU}|{expr}}$ is defined deterministically from Table \ref{relationship},
and is 1 for prototypical or observational AUs, and 0 otherwise. 

Then we match MMA's AU representations ($r_{AU}$) with the AU pseudo-representations by minimizing the binary cross entropy with soft targets loss:

\begin{equation}
    \mathcal{L}_{DM} = \mathbb{E} \Bigg[ 
    \sum_{i=1}^{17}[ - {r^\prime}_{AU}^{\text{ }i} \cdot \text{log 
    }   r_{AU}^i
    ] \Bigg]
\label{eq:distr}
\end{equation}

With this loss we aim to infuse prior knowledge on task's relationship (according to Table \ref{relationship}) into the network, so as to guide the generation of better and consistent expression and AU representations.
For instance, if the network predicts \emph{happiness} with probability 1 and also predicts that AUs 4, 15 and 1 are activated, this is a mistake as these AUs are  associated with the expression \emph{sadness}. In this case, the AU and expression representations are in conflict.
Therefore the overall objective function ($\mathcal{L}_{MMA}$) minimized during MMA's training is:

\begin{align}
\mathcal{L}_{MMA} = \mathcal{L}_{CCC} + \mathcal{L}_{CCE} + \mathcal{L}_{BCE} + \mathcal{L}_{DM} \label{eq:coupleloss}
\end{align}

\noindent where: \\
$\mathcal{L}_{CCC}$ is the loss term associated with the valence-arousal estimation task, and $\mathcal{L}_{CCC} = 1 - 0.5 \cdot (\rho_a + \rho_v)$, with $\rho_{a/v}$ being the Concordance Correlation Coefficient (CCC) of arousal/valence;  \\
$\mathcal{L}_{CCE}$ is the categorical cross entropy loss associated with the 7 basic expression recognition task; and \\
$\mathcal{L}_{BCE}$ is the binary cross entropy loss associated with the 17 AU detection task.

\subsection{MRNN: Masked RNN and Routing Component}

As described in the previous section, the MMA component processes an input video $\textbf{\textit{X}}$ by extracting affective representations from each frame ($\textbf{\textit{x}}_i$) using three distinct affective models. This results in an affect-level representation matrix $\textbf{\textit{Z}} = \{\textbf{z}_1,..., \textbf{z}_t\} \in \Re^{26 \times t}$. 
This matrix is then fed into an RNN positioned atop the MMA component, which captures temporal dependencies and sequential information across consecutive frames of the video. The RNN sequentially processes the extracted vector representations from frame $0$ to frame $t$, mapping these representations 
$\{\textbf{z}_k\}^K_{k=1}$ to embeddings $\{\textbf{z}'_k\}^K_{k=1}$, where each $\textbf{z}'_k \in \Re^{d'}$.

These embeddings (corresponding to all video frames) are concatenated into a single vector embedding $\textbf{\textit{z}}' \in \Re^{d' \cdot t}$, aligning with the goal of estimating the intensity of various facial expressions across the entire video sequence, consistent with the provided annotations. This embedding $\textbf{\textit{z}}'$ is then passed through a Mask layer, producing new embedding $\textbf{\textit{z}}'' \in \Re^{d' \cdot t}$. The original (pre-padding) length $l$ of the input video is propagated to the Mask layer to guide the routing mechanism. During training, the routing mechanism dynamically selects elements from various positions within $\textbf{\textit{z}}'$ based on the video's length $l$, preserving the values of these selected elements and setting the remaining elements to zero. This process effectively routes only the relevant elements into the subsequent layer, thereby enhancing the model's focus on key temporal features.

The embedding $\textbf{\textit{z}}'' \in \Re^{d' \cdot t}$ is then transformed into another embedding $\textbf{\textit{z}}''' \in \Re^{d''}$ through a feed forward layer,  which is trained to extract high-level information from the 'masked'  embedding $\textbf{\textit{z}}''$.
During training, only the weights connecting the feed-forward layer neurons to the elements within $\textbf{\textit{z}}'$ routed by the Mask layer are updated. The remaining weights are updated when their corresponding feed-forward layer neurons are connected to elements within $\textbf{\textit{z}}'$ that are selected by the Mask layer in a different video input. The loss function minimization is conducted similarly to networks with dynamic routing, where the weights not involved in the routing process remain constant, and links corresponding to non-routed elements within $\textbf{\textit{z}}'$ are ignored.
Finally, the embedding $\textbf{\textit{z}}'''$ is mappes to the logits $\textbf{\textit{u}}$ via a feed forward layer parameterized by $\textbf{\textit{W}}$, resulting in the video-level classification, expressed as: $ 
 \textbf{\textit{u}} =  \textbf{\textit{W}}^T \textbf{\textit{z}}'''$.

The loss function that we utilized for training MMA-MRNNet:  was not the typical Mean Squared Error (MSE) but a loss based on the pearson correlation since that correlation metric was the evaluation criterion for the utilized database:

\begin{equation} \label{eq:3}
\mathcal{L}_{total} = 1 - \sum_{i=1}^{7} \frac{\rho_i} {7} = 1 - \frac{1}{7} \sum_{i=1}^{7} \frac{s_{i,xy}}{\sqrt{s_{i,x} \cdot s_{i,y}}}
\end{equation}

\noindent
where: $i$ denotes the facial expression; $\rho_i$ is the pearson correlation coefficient; $s_{i,x}$ and $s_{i,y}$ are the variances of the expression labels and predicted values; $s_{i,xy}$ is their covariance.

\subsection{Datasets, Pre-Processing and Implementation Details}

The Hume-Reaction dataset was used as part of both the Emotional Reactions Sub-Challenge of MuSe 2022 \cite{Christ22-TM2} and the Emotional Reaction Intensity Estimation Challenge of the 5th ABAW Competition in 2023 \cite{kollias2019expression,kollias2020analysing,kollias2021affect,kollias2021analysing,kollias2021distribution,kollias2020deep,zafeiriou1,kollias2017recognition,kollias20222abaw,kollias2022abaw,kollias2023abaw2,kollias2023deep2,kollias2023abaw,kollias2023abaww,kollias2023facernet,kollias2023multi,kollias20246th,kollias2024distribution,kollias2024domain,arsenos2023data,kollias2023ai}. The participants of this subchallenge explore a multi-output regression task, utilizing seven, self-annotated, nuanced classes of emotion: ‘Adoration,’ ‘Amusement,’ ‘Anxiety,’ ‘Disgust,’ ‘Empathic-Pain,’
‘Fear,’ and ‘Surprise.’ The dataset consists of 25,067 videos taken from 2,222 subjects of which 15,806 constitute the training set, 4,657 the validation set and 4,604 the test set. 

The Aff-Wild2 database \cite{kollias2019deep,kollias2019expression,kollias2022abaw,kollias2023abaww,kollias2023multi,kollias20246th,kollias2024distribution,kolliasijcv,zafeiriou2017aff,kollias2020analysing,kollias2021distribution,kollias2021affect,kollias2019face,kollias2021analysing,kollias20247th} is the largest in-the-wild database and the only one to be annotated in a per-frame basis for the seven basic expressions (i.e., happiness, surprise, anger, disgust, fear, sadness and the neutral state), twelve action units (AUs 1,2,4,6,7,10,12,15,23,24,25, 26) and valence and arousal. In total, it consists of 564 videos of around 2.8M frames with 554 subjects. Aff-Wild2 displayes a big diversity in terms of subjects' ages, ethnicities and nationalities; it has also great variations and diversities of environments.

The AffectNet dataset \cite{mollahosseini2017affectnet} contains around 1M facial images, 300K of which were manually annotated in terms of 7 discrete expressions (plus contempt) and valence-arousal.
The original training set of this database consists of around 290K images and the original validation of 4K. We evaluate our method on the updated partitioning protocol of this database according to our previous work \cite{hu2024bridging,hu2024rethinking} (as we mentioned in Section 3.1). This new partitioning consists of a training set of around 160K images, a validation set of around 45K images and a test set of around 90K images.

The EmotioNet database \cite{emotionet2016} contains around 1M images and was released for the EmotioNet Challenge in 2017. 950K images were automatically annotated and the remaining 50K images were manually annotated with 11 AUs (1,2,4,5,6,9,12,17,20,25,26); around half of the latter constituted the validation and the other half the test set of the Challenge. 
We evaluate our method on the updated partitioning protocol of this database according to our previous work \cite{hu2024bridging,hu2024rethinking} (as we mentioned in Section 3.1). This new partitioning consists of a training set of around 25K images, a validation set of around 7K images and a test set of around 14K images.

We used the RetinaFace detector \cite{deng2020retinaface} to extract, from all images, face bounding boxes and 5 facial landmarks; the latter were used  for face alignment. All   cropped   and   aligned   images   were  resized   to $112 \times 112 \times 3$ pixel resolution and their intensity values were normalized  to  $[-1,1]$.

We chose batch size equal to 4, length $t$ equal to 480, Adam optimizer with learning rate $10^{-4}$ when training from scratch and $10^{-5}$ when training in an end-to-end manner, after having initialised each subnetwork. For RNN we utilize 1-layer GRU with 128 units; feed forward layer consists of 32 units.
Training was performed on a Tesla V100 32GB GPU; training time was 3 days. The TensorFlow platform has been used. 


\section{Experimental Results}


\subsection{Comparison with the state-of-the-art}

At first we compare the performance of MMA-MRNNet to that of various baseline \cite{Christ22-TM2} and state-of-the-art methods: ViPER and Netease Fuxi Virtual Human methods (which are multi-modal methods exploiting audio, visual and text information); the best performing HFUT-CVers method (presented in the related work section; it is an ensemble multi-modal method exploiting both audio and visual information); USTC-IAT-United method (which was presented in the related work section and is a multi-modal method exploiting both audio and visual information); USTC-AC and NISL-2023 methods (both presented in the related work section; they are ensemble multi-modal methods exploiting both audio and visual information). 
Table \ref{aaaa} shows that our uni-modal non-ensemble learning MMA-MRNNet (that exploits only the visual information and does not employ any ensemble learning) outperforms all other methods by large margins (although some methods are multimodal ones or even ensembles). Let us also note that all baseline and state-of-the-art methods utilized the ad-hoc strategy of selecting fixed input length by removing or duplicating images within each video sequence.

\begin{table}[h]
\caption{Comparison between MMA-MRNNet, baselines and the state-of-the-art on the test set of Hume-Reaction dataset; Pearson’s Correlation Coefficient results are denoted in \%}
\centering
\scalebox{1.1}{
\begin{tabular}{cc}
\hline
Methods                                 & Pearson’s Correlation Coefficient ($\rho$)

\\ \hline \hline

HFUT-CVers \cite{li2023multimodal}
&	47.3
\\
USTC-IAT-United   \cite{yu2023exploring}
& 43.8
\\ 
Netease Fuxi Virtual Human \cite{qiu2023multi} & 40.5 \\
USTC-AC \cite{wang2023emotional}	&	37.3  \\
NISL-2023 \cite{zhang2023facial}	&	36.7
\\
ViPER \cite{vaiani2022viper}  & 29.7      \\ 
FAU-Baseline \cite{Christ22-TM2} & 28.0 \\
VGGface 2-Baseline \cite{Christ22-TM2} & 18.3 \\
Fusion-Baseline \cite{Christ22-TM2} & 20.30 \\ \hline
\textbf{MMA-MRNNet} & \textbf{53.1}      \\ \hline
\end{tabular}
}
\label{aaaa}
\end{table}

\subsection{Ablation Study}

We conducted a series of ablation experiments to evaluate the impact of different elements and components on our model's performance. 

Initially, we used only single-task affective representations (extracted from MMA) as input to the RNN. We then tested combinations of two tasks (e.g., VA \& AUs), and finally, we utilized the affective representations from all three tasks concurrently. The results are summarized in Table \ref{tab2}, where we present only the best performance for each experiment to avoid cluttering of the results. Notably, even when using only valence and arousal representations, our network outperformed all other methods except HFUT-CVers. The model's performance improved substantially when incorporating additional per-frame features, such as the 7 basic expressions or 17 AUs. In the two-task experiments, we observed a further increase in the Pearson's correlation coefficient ranging from 1\% to 1.5\%. Ultimately, when all three tasks were used together, our method achieved the highest performance.

\begin{table}[h]
\caption{Ablation Results on MMA-MRNNet on the validation set of Hume-Reaction dataset; Pearson’s Correlation Coefficient results are denoted in \%}
\centering
\scalebox{1.}{
\begin{tabular}{cc}
\hline
\begin{tabular}{@{}c@{}} \textbf{Affective Representation} \\ \textbf{ from the MMA component}\end{tabular}  
                             & \textbf{Pearson’s Correlation Coefficient ($\rho$)} \\ \hline \hline
VA                                       & 51.1                                           \\ 
7 Basic Expressions                                    & 52.6                                           \\ 
17 AUs                                         & 53.0                                           \\ 
VA \& 7 Basic Expressions                 & 53.5                                           \\ 
VA \& 17 AUs                        & 54.1                                           \\ 
17 AUs \& 7 Basic Expressions                    & 54.1                                           \\ \hline
\textbf{VA \& 7 Basic Expressions \& 17 AUs} & \textbf{54.4}                                           \\ \hline
\end{tabular}
}
\label{tab2}
\end{table}

To identify the optimal architecture for our network, we conducted experiments with various configurations, including different CNNs (e.g., ResNet50 instead of MMA) and RNNs (e.g., LSTM instead of GRU), as well as varying the number of layers and units, as detailed in Table \ref{tab3}. After evaluating a wide range of combinations, we determined that the most effective configuration consists of a single GRU layer with 128 units, followed by a feed forward layer with 32 units. Additionally, we evaluated the impact of incorporating the Mask layer, dynamic routing, and our proposed loss function (as an alternative to the conventional MSE). The results presented in Table \ref{tab2} demonstrate that these components significantly enhance the performance of MMA-MRNNet.

\begin{table}[h]
\caption{Further Ablation Results on MMA-MRNNet on the validation set of Hume-
Reaction dataset; Pearson’s Correlation Coefficient results are denoted in \%}
\centering
\scalebox{1.}{
\begin{tabular}{cc}
\hline
\textbf{Model}                                 & Pearson’s Correlation Coefficient ($\rho$)

\\ \hline \hline
MMA + GRU + FC (64)                                    & 53.0                                           \\ 
MMA + GRU + FC (16)                                    & 53.5                                           \\                
MMA + GRU + FC (8)                                    & 52.8                                           \\ 
MMA + 2 $\times$ GRU + FC (32)                                    & 54.2         \\
ResNet50 + GRU + FC (32)                                    & 51.5     \\ 
MMA + LSTM + FC (32)                                    & 53.2     \\
MMA + GRU (256) + FC (32)                                    & 53.4     \\
MMA + GRU (64) + FC (32)                                    & 53.8     \\
\hline
MMA-MRNNet  w/o Mask \& Routing                                & 51.9                                           \\ 
MMA-MRNNet with MSE                                   & 53.1                                           \\ 
\hline
\textbf{MMA-MRNNet}                                   & \textbf{54.4}                                           \\ 

\hline
\end{tabular}
}
\label{tab3}
\end{table}


\subsection{MMA Evaluation Results}

Here we provide an extensive experimental study in which we utilise the top-performing methods from the ABAW Competitions 
(FUXI \cite{fuxi}, SITU \cite{situ}, CTC \cite{ctc}) and the state-of-the-art  methods (DACL \cite{dacl}, DAN \cite{dan}, POSTER++ \cite{poster}; ME-GraphAU \cite{megraphau} \& AUNets \cite{aunets})
for 7 basic expression recognition, AU detection and valence-arousal estimation, and compared their performance to our proposed MMA component.

As can be seen on Table \ref{comparison_sota_mma}, our proposed MMA component outperformed all these methods on all tasks (7 basic expression recognition, AU detection and valence-arousal estimation) and on all utilized databases (Aff-Wild2, AffectNet and EmotioNet) by large margins.

\begin{table*}[h]
\caption{
Performance comparison (in \%) between the MMA component and various state-of-the-art methods. 'CCC-VA' represents the average Concordance Correlation Coefficient (CCC) for valence and arousal. 'F1 - Expr' refers to the average (i.e., macro) F1 score across the 7 basic expressions, while 'F1 - AUs' denotes the average F1 score across all AUs present in each of the databases used.
}
\centering
\begin{tabular}{ c||c|c|c||c|c||c }
 \hline
\multicolumn{1}{c||}{\begin{tabular}{@{}c@{}} Databases  \end{tabular}}  &  \multicolumn{3}{c||}{\begin{tabular}{@{}c@{}}  Aff-Wild2 \end{tabular}} & \multicolumn{2}{c||}{\begin{tabular}{@{}c@{}}  AffectNet \end{tabular}} & \multicolumn{1}{c}{\begin{tabular}{@{}c@{}}   EmotioNet\end{tabular}}      \\
 \hline
Methods & CCC-VA & F1 - Exprs &  F1 - AUs & CCC-VA & F1 - Exprs & F1 - AUs \\
 \hline
 \hline

\begin{tabular}{@{}c@{}}  SITU \end{tabular}  &  64.14  & 38.24  & 54.22  & 71.1  & 59.0  & 77.2   \\
\hline

\begin{tabular}{@{}c@{}}  CTC \end{tabular}  & 56.66   & 33.11  & 48.87  & 71.0  & 57.7 &  74.6  \\
\hline

\begin{tabular}{@{}c@{}}  FUXI \end{tabular}  &  63.72  &  39.21  & 55.49  & 74.0  &  63.1 &  77.9  \\
\hline

\begin{tabular}{@{}c@{}}  ME-GraphAU \end{tabular}  & -   & -  & -  & -  & - &   72.9 \\
\hline

\begin{tabular}{@{}c@{}}  AUNets \end{tabular}  & -   & -  & -  & -  & - &  82.8  \\
\hline

\begin{tabular}{@{}c@{}}  DAN \end{tabular}  &  -  & -  & -  & -  & 60.0 &  -  \\
\hline

\begin{tabular}{@{}c@{}}  DACL \end{tabular}  &  -  & -  & -  & -  & 60.3 &  -  \\
\hline

\begin{tabular}{@{}c@{}}  POSTER++ \end{tabular}  &  -  & - & -  & -  & 63.2 & -   \\
\hline
\hline

\begin{tabular}{@{}c@{}}  \textbf{MMA} \end{tabular}  &  \textbf{67.38}  & \textbf{43.21}  &  \textbf{58.87}  & \textbf{78.2}  & \textbf{65.4} &  \textbf{85.4}  \\
\hline

\end{tabular} 
\label{comparison_sota_mma}
\end{table*}

The performance of the proposed MMA method was evaluated against several state-of-the-art approaches across multiple datasets (Aff-Wild2, AffectNet and EmotioNet), as detailed in Table \ref{comparison_sota_mma}. The MMA component consistently outperformed by large margins all methods on all tasks (7 basic expression recognition, AU detection and valence-arousal estimation) and on all utilized databases, across all evaluation metrics. 
Specifically, on the Aff-Wild2 dataset, MMA surpassed the closest competitor, SITU, for valence-arousal estimation, by 3.24\%. It also outperformed the closest competitor, FUXI, for 7 basic expression recognition by 4\%, as well as for AU detection by 3.38\%. On the AffectNet dataset, MMA again demonstrated superior performance, outperforming the sota methods by at least 4.2\% for valence-arousal estimation and by at least 2.2\% for 7 basic expression recognition. On the EmotioNet dataset, MMA outperformed the sota methods by at least 2.6\%. These results underscore the robustness and superiority of MMA in delivering precise and reliable affect representations.

\section{Conclusion}

In this paper, we introduced MMA-MRNNet, a novel deep learning architecture for dynamic multi-output Facial Expression Intensity Estimation (FEIE) from video data. Our method addresses the limitations of traditional approaches by leveraging a Multi-Task Learning (MTL) framework to extract rich affective representations, including valence-arousal, basic facial expressions, and action units (AUs). These representations are further refined through a Masked Routed RNN (MRNN), which dynamically adjusts to the variable lengths of input videos, ensuring robust and accurate predictions.

We demonstrated the effectiveness of MMA-MRNNet on the Hume-Reaction dataset, where it consistently outperformed by large margins all state-of-the-art methods. We also demonstrated the effectiveness of the MMA component across multiple in-the-wild datasets, where it consistently outperformed all state-of-the-art methods across various metrics.
Our approach not only handles the complexities of video-level annotation but also mitigates the challenges associated with processing variable-length sequences, offering a flexible and powerful solution for real-world applications in affective computing.

\bibliographystyle{splncs04}
\bibliography{main}
\end{document}